\documentclass{article}
\usepackage{ijcai22}

\usepackage{times}
\usepackage{soul}
\usepackage{url}
\usepackage{hyperref}
\usepackage[utf8]{inputenc}
\usepackage[small]{caption}
\usepackage{graphicx}
\usepackage{amsmath}
\usepackage{amsthm}
\usepackage{amssymb,amsfonts}
\usepackage{booktabs}
\usepackage{algorithm}
\usepackage{algorithmic}
\title{Constrained Reinforcement Learning for Dexterous Manipulation}

\author{
Abhineet Jain\and
Jack Kolb\And
Harish Ravichandar
\affiliations
Georgia Institute of Technology\\
\emails
\{abhineetjain, kolb\}@gatech.edu,
harish.ravichandar@cc.gatech.edu}

\begin{document}

\maketitle

\begin{abstract}
    Existing learning approaches to dexterous manipulation use demonstrations or interactions with the environment to train black-box neural networks that provide little control over how the robot learns the skills or how it would perform post training. These approaches pose significant challenges when implemented on physical platforms given that, during initial stages of training, the robot's behavior could be erratic and potentially harmful to its own hardware, the environment, or any humans in the vicinity.
    A potential way to address these limitations is to add constraints during learning that restrict and guide the robot's behavior during training as well as roll outs. Inspired by the success of constrained approaches in other domains, we investigate the effects of adding position-based constraints to a 24-DOF robot hand learning to perform object relocation using Constrained Policy Optimization. We find that a simple geometric constraint
    can ensure the robot learns to move towards the object sooner than without constraints. 
    Further, training with this constraint
    requires a similar number of samples as its unconstrained counterpart to master the skill. These findings shed light on how simple constraints can help robots achieve sensible and safe behavior quickly and ease concerns surrounding hardware deployment. We also investigate the effects of the strictness of these constraints and report findings that provide insights into how different degrees of strictness affect learning outcomes. Our code is available at \textit{\url{https://github.com/GT-STAR-Lab/constrained-rl-dexterous-manipulation}}.
\end{abstract}

\begin{figure}[t]
    \centering
    \includegraphics[width=0.35\textwidth]{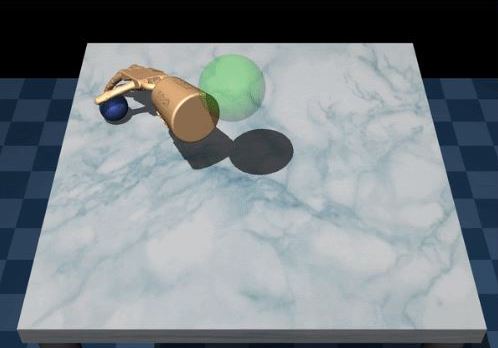}
    \caption{The relocation task in MuJoCo. This task requires the robot hand to pick up the blue ball from the tabletop and carry it to the green goal region.}
    \label{fig:task}
\end{figure}

\section{Introduction}

Dexterous manipulation often involves the use of high degree-of-freedom robots to manipulate objects. Representative dexterous manipulation tasks include relocating objects, picking up arbitrarily shaped objects, and sequential interactions with articulated objects (e.g. unlatching and opening a door). Indeed, factors such as high-dimensional state spaces and complex interaction dynamics make these tasks challenging to automate. Classical control methods
are hard to recruit for dexterous manipulation due to the manual effort required to design controllers in high-dimensional spaces.

Prior work in dexterous manipulation has succeeded by using self-supervised methods in simulation, and transferring learned policies to real robots \cite{akkaya2019solving}. Others have utilized demonstrations to improve reinforcement learning \cite{rajeswaran2017learning}. However, these approaches are hard to train on real robots
, as initial robot behavior can be erratic and unsafe.



\begin{figure}[t]
    \centering
    \includegraphics[width=0.35\textwidth]{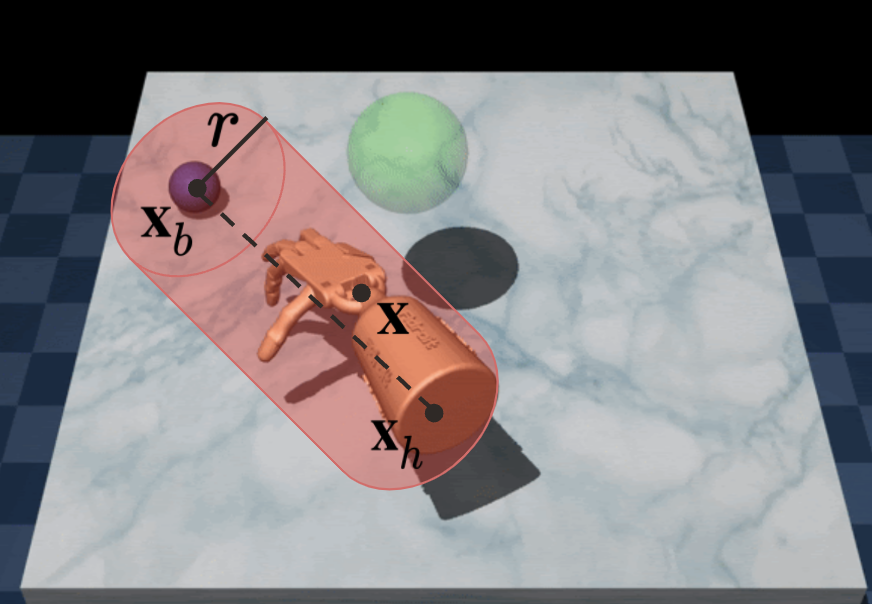}
    \caption{The boundary constraint defined in Eq. \ref{eq:constraint} for the initial motion of the robot hand towards the ball. The robot incurs a penalty when it moves outside the boundary.}
    \label{fig:taskconstraint}
\end{figure}

In this work, we explore adding instance-specific constraints to an object relocation task (Fig. \ref{fig:task}), that restrict and guide the robot's behavior during training as well as roll outs.
Constrained Policy Optimization (CPO) is an effective method to solve constrained MDPs \cite{achiam2017constrained}, built upon trust-region policy optimization (TRPO) \cite{DBLP:journals/corr/SchulmanLMJA15}. We formulate a cylindrical boundary constraint for the initial motion of the robot hand towards the object (Fig. \ref{fig:taskconstraint}). The robot incurs a penalty when it moves outside the boundary. We find that using CPO with this simple geometric constraint can ensure the robot learns to move towards the object sooner than without constraints. Further, training with this constraint (CPO) requires a similar number of samples as its unconstrained counterpart (TRPO) to master the skill. These findings shed light on how simple constraints can help robots achieve sensible and safe behavior quickly and ease concerns surrounding hardware deployment. We also investigate the effects of the strictness of these constraints and report findings that provide insights into how different degrees of strictness affect learning outcomes.


\begin{figure*}[t]
    \centering
    \includegraphics[width=1.0\textwidth]{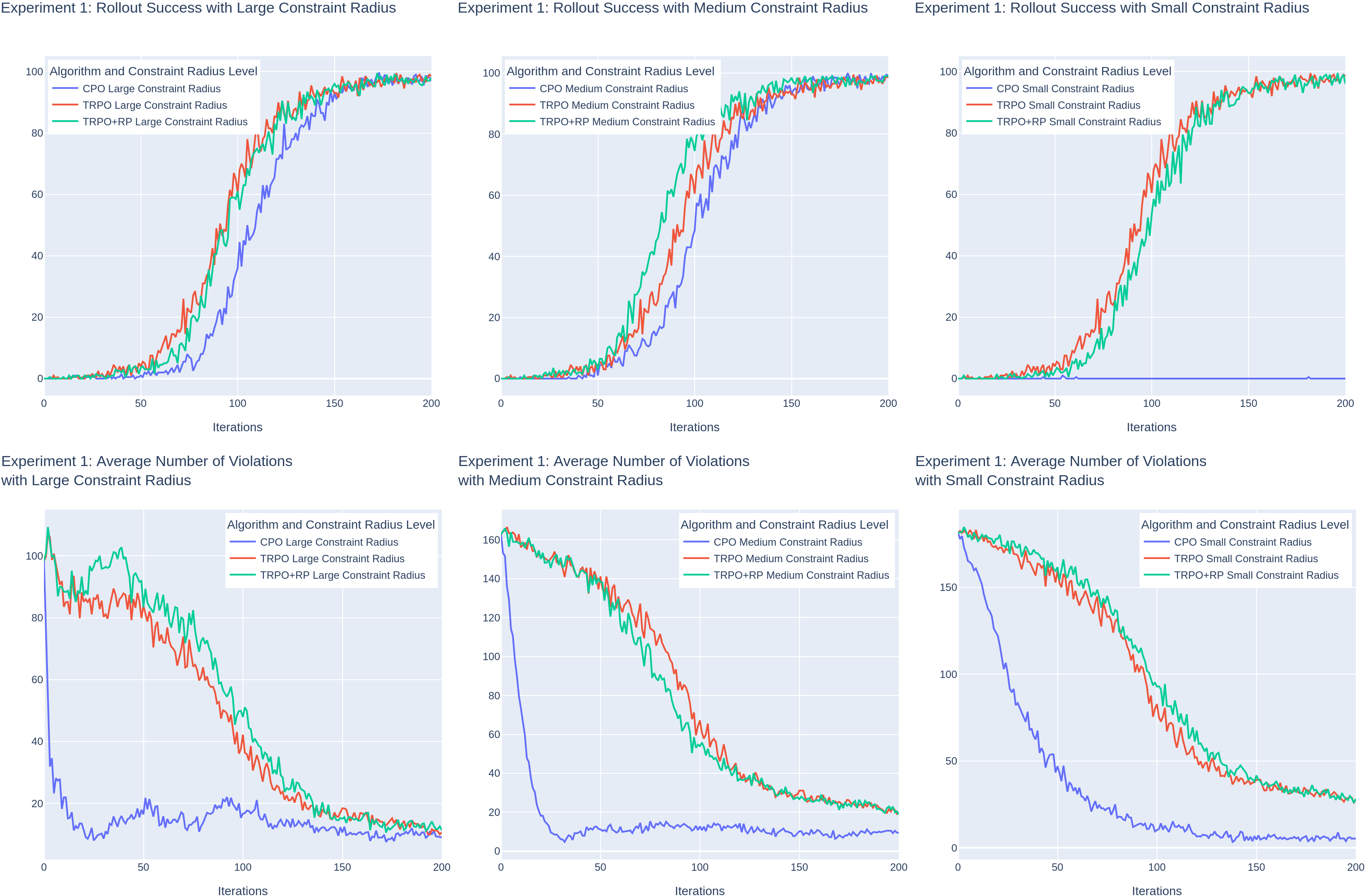}
    \caption{\textbf{Comparing CPO with TRPO and TRPO-RP}. \textbf{Bottom}: CPO has \textit{reduced average number of violations} than both the TRPO policies for all the three different intensities of the boundary constraint. CPO learns to satisfy the constraints better throughout the training process. \textbf{Top}: CPO has \textit{lower sample efficiency} for all the three constraint cases, which is a small trade-off to ensure safe learning.}
    \label{fig:exp1detailed}
\end{figure*}

\section{Related Work}

Previous works explore self-supervised methods to manipulate objects by adding different types of constraints. To gently lift objects, tactile sensors have been used to constrain contact forces in a 24-DOF robot hand \cite{huang2019learning}. However, this approach does not consider task performance. Another work trains dynamic motion primitives (DMP) for a 10-DOF robot hand considering virtual joint constraints or friction \cite{li2017reinforcement}. This approach does not provide any safety guarantees beyond DMPs being deterministic. In low-dimensional environments, one work looks into stable robot path trajectories using graph optimization for motion planning, but does not focus on performance \cite{englert2016combined}. Another work focuses on in-hand object manipulation in a low-dimensional environment by adding constraints between the robot and object \cite{kobayashi2005reinforcement}. In multi-agent settings, boundaries based on robot geometry have been used to enable safe collaboration in close quarters to provide safety guarantees \cite{gu2017deep}.

\section{Background}

\subsection{Trust Region Policy Optimization (TRPO)}

TRPO is a policy gradient method to solve Markov Decision Processes, that avoids parameter updates that change the policy too much with a KL divergence constraint on the size of the policy update at each iteration \cite{DBLP:journals/corr/SchulmanLMJA15}.

We train TRPO on-policy i.e., the policy for collecting data is same as the policy that we want to optimize. The objective function $J(\theta)$  measures the total advantage $\hat A_{\theta_{old}}$ over the state visitation distribution $p^{\pi_{\theta_{old}}}$ and actions from $\pi_{\theta_{old}}$, while the mismatch between the training data distribution and the true policy state distribution is compensated with an importance sampling estimator. TRPO aims to maximize the objective function subject to a trust region constraint which enforces the distance between old and new policies measured by KL-divergence to be small enough, within a parameter $\delta$. The same can be summarized in Eq. \ref{eq:trpo}.

\begin{equation}
\label{eq:trpo}
\resizebox{.91\linewidth}{!}{$
            \displaystyle
            \begin{aligned}
            \text{maximize}\ J(\theta) = \mathbb{E}_{s \sim p^{\pi_{\theta_{old}}}, a \sim \pi_{\theta_{old}}}\left(\frac{\pi_\theta(a | s)}{\pi_{\theta_{old}}(a | s)} \hat A_{\theta_{old}}(s, a)\right)
            \\
            \text{subject to}\ \mathbb{E}_{s \sim p^{\pi_{\theta_{old}}}}\left[D_{KL}(\pi_{\theta_{old}}(. | s) \parallel \pi_{\theta}(. | s)\right] \leq \delta
            \end{aligned}
        $}
\end{equation}

\subsection{Constraint Policy Optimization (CPO)}

CPO \cite{achiam2017constrained} is built on top of the TRPO algorithm to solve Constrained Markov Decision Processes (CMDPs) which include a cost function, $C$ and a cost discount factor $\gamma_c$ along with the standard MDP learning problem $(S, A, T , R, p_0, \gamma)$. In a local policy search for CMDPs, on top of the TRPO optimization, we additionally require policy updates to be feasible for the CMDP. Our objective function thus adds another condition to limit the expected discounted cost under a cost limit, $cl$ for each constraint. The same can be summarized in Eq. \ref{eq:cpo}.

\begin{equation}
\label{eq:cpo}
    \resizebox{.91\linewidth}{!}{$
            \displaystyle
            \begin{aligned}
            \text{maximize}\ J(\theta) = \mathbb{E}_{s \sim p^{\pi_{\theta_{old}}}, a \sim \pi_{\theta_{old}}}\left(\frac{\pi_\theta(a | s)}{\pi_{\theta_{old}}(a | s)} \hat A_{\theta_{old}}(s, a)\right)
            \\
            \text{subject to}\ \mathbb{E}_{s \sim p^{\pi_{\theta_{old}}}}\left[D_{KL}(\pi_{\theta_{old}}(. | s) \parallel \pi_{\theta}(. | s)\right] \leq \delta
            \\
            \text{and}\ \mathbb{E}_{\tau \sim \pi_{\theta}}\left[\sum_{t=0}^{\infty}\gamma_c^tC_i(s_t, a_t, s_{t+1})\right] \leq cl_i \forall i
            \end{aligned}
        $}
\end{equation}


\subsection{Problem Formulation}

We consider an object relocation task where the agent, a 24-DOF Adroit hand, learns a policy to grasp and relocate a blue ball from a tabletop to a green region (Fig. \ref{fig:task}). We formulate this task as a Constrained MDP, $(S, A, T , R, p_0, \gamma, C, \gamma_c)$, where $S$ is the state space, $A$ is the action space, $T$ is the transition function, $R$ is the reward function, $p_0$ is the initial state distribution, $\gamma$ is a discount factor, $C$ is the cost function and $\gamma_c$ is a cost discount factor.

In a typical episode, at each time $t$, the agent
receives an observation $o_t$ based on the current state. After the agent takes an action $a_t \sim \pi_{\theta}(o_t)$ based on the observation, it gets reward $R(s_t)$ from the environment, incurs a penalty cost $C(s_t)$ and arrives at a new state with observation $s_{t+1} = T(s_t, a_t)$. Based on the RL algorithm, CPO or TRPO, the agent optimizes the corresponding objective function. 

\textbf{Observation space}: The observation space is 39-dimensional including 24 hand joint angles, hand translation (3-D), hand orientation (3-D), relative position of the hand with respect to the object (3-D), relative position of the hand with respect to the goal (3-D) and relative position of the object with respect to the goal (3-D).

\textbf{Action space}: Each action for the relocation task is 30-dimensional, including 24 hand joint angles, 3-D hand translation and 3-D hand rotation.

\textbf{Reward}: The agent is rewarded for getting closer to the object, lifting the object up, taking the object closer to the goal, taking the hand closer to the goal, and reaching really close to the goal. 

\subsection{Constraint Formulation}
We define boundary constraints to restrict the initial motion of the robot arm in the direction of the object for relocation. Considering $\textbf{x}_h$ as the initial hand position, $\textbf{x}_b$ as the initial object position, $\textbf{x}$ as the current hand position, the constraints are defined in Eq. \ref{eq:constraint}, where $r$ is the boundary radius. The same can be visualized in Fig. \ref{fig:taskconstraint}. The derivation of the constraints can be found in Appendix \ref{appendix:a}.

\begin{equation}\label{eq:constraint}
  \begin{aligned}
\frac{|(\textbf{x}_b - \textbf{x}_h) \times (\textbf{x}_h - \textbf{x})|}{|\textbf{x}_b - \textbf{x}_h|} \leq r
\\
0 \leq  \frac{(\textbf{x} - \textbf{x}_h)\ .\ (\textbf{x}_b - \textbf{x}_h)}{|\textbf{x}_b - \textbf{x}_h|^2} \leq 1
 \end{aligned}
\end{equation}

We penalize the agent with a fixed penalty cost whenever the agent violates any of the formulated constraints. If it violates both, it receives twice the penalty cost (set to 0.01). For practical purposes, we relax the second constraint to range between -0.1 and 1.1. A tight second constraint does not allow the robot to learn grasping the ball.

\section{Experiments}

\begin{figure*}[t]
    \centering
    \includegraphics[width=1.0\textwidth]{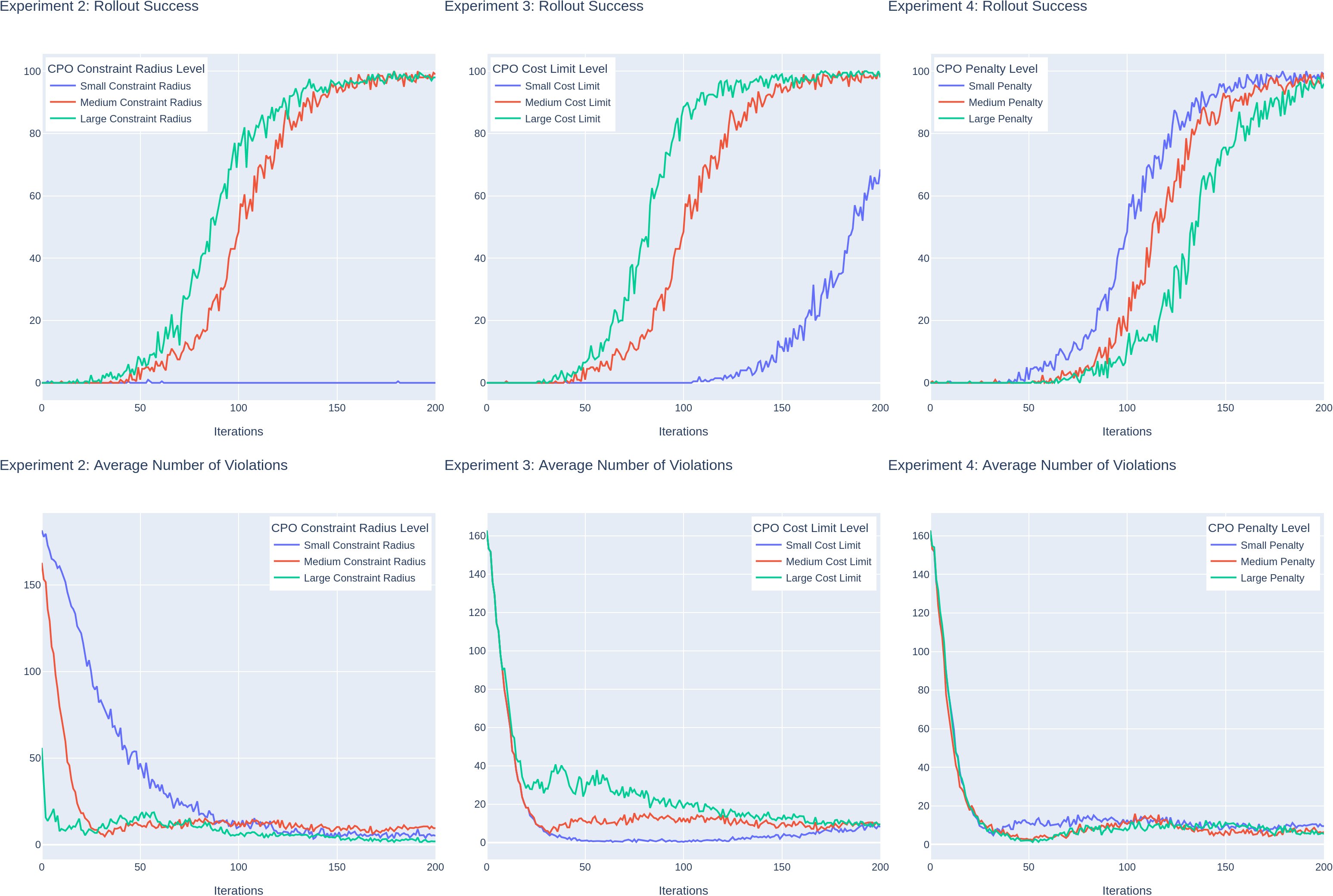}
    \caption{\textbf{Experiment with constraint parameters.} \textbf{Left}: (Top) A smaller radius takes significantly more samples to train, whereas relaxed radii are more sample efficient. (Bottom) The average number of violations also reduce more quickly for a relaxed constraint than a smaller one, although they are obviously lower to begin with. \textbf{Center}: (Top) As the cost limit decreases, the sample efficiency also decreases. (Bottom) CPO optimizes perfectly for the respective limits and maintains the allowed cost throughout most of the training. \textbf{Right}: (Top) Higher the penalty cost for each violation, the longer it takes for the policy to train. (Bottom) Scaling of the penalty costs does not really impact how the average number of violations reduce during training. }
    \label{fig:exp234}
\end{figure*}

\begin{figure}[t]
    \centering
    \includegraphics[width=0.45\textwidth]{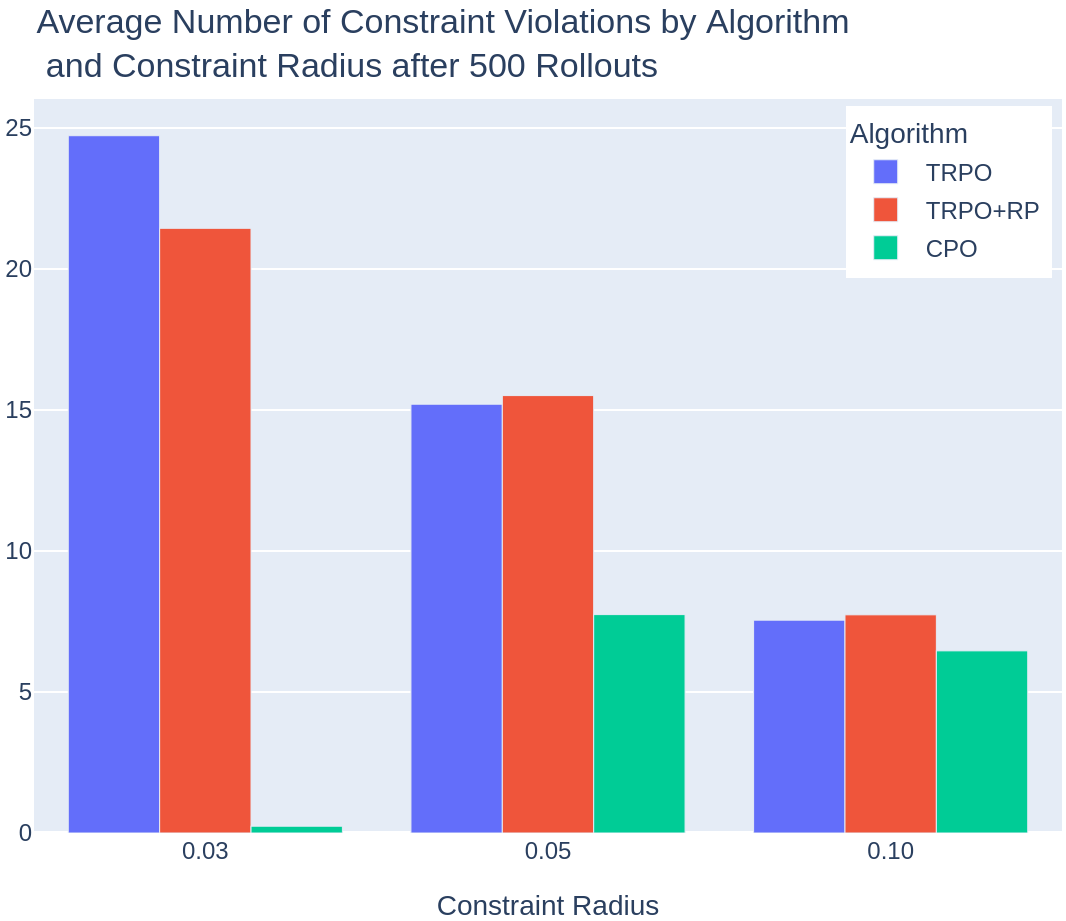}
    \caption{\textbf{Average constraint violations on 500 roll outs after training.} The CPO policy maintains lesser number of violations and is thus safer to roll out in the real world than the TRPO policies.}
    \label{fig:testbargraph}
\end{figure}

We design different experiments to evaluate the following research questions: 
\begin{enumerate}
    \item Does the policy learned via CPO allow \textit{safe training} and \textit{safe roll outs}? How does it compare to a TRPO policy?
    \item What is the effect of changing the constraint \textit{boundary radius} on training CPO?
    \item What is the effect of changing the allowed \textit{cost limit} on training CPO?
    \item What is the effect of changing the \textit{cost of each penalty} on training CPO?
\end{enumerate}


\subsection{Setup}

We use the MuJoCo physics simulator as our testing platform. All our RL policies are pre-trained using behavior cloning via 25 demonstrations collected via CyberGlove III \cite{rajeswaran2017learning}. Our policy network is a Gaussian Multi-Layer Perceptron (MLP) with two hidden layers of 32 neurons each. We also train a value network and a cost value network (only for CPO), both MLPs with two hidden layers of 128 neurons each. The learning rate for behavior cloning on our policy network, value network and cost value network is 0.001. For training via CPO and TRPO algorithms, our reward and cost discount factors are both 0.995 and the GAE parameter is 0.97. The constraint configurations for our different experiments are detailed in Appendix \ref{appendix:b}.

\subsection{Experiment 1: Evaluating CPO and TRPO}

We evaluate both CPO and TRPO algorithms on the relocation task to verify the effect of optimizing for constraints on sample efficiency and average cost during training. We evaluate two variants of the TRPO algorithm -- one where the reward is penalized with the incurred cost (TRPO-RP), and one without the penalty (TRPO).

We see that CPO has a \textbf{lower average number of violations} than both the TRPO policies for three different intensities of the boundary constraint. From Fig. \ref{fig:exp1detailed} (bottom), we see that CPO learns to satisfy the constraints better throughout the training process, making it potentially safer to train on real robots. Qualitative results showing the behavior for all three algorithms during early, mid and late training can be found in the video at \href{https://drive.google.com/file/d/1ChOgY2DQdouAOw1s3CkKZoj0yQywGxfj/view?usp=sharing}{\textit{this link}}. CPO also has \textbf{lower sample efficiency} for the three constraint cases, which is a small trade-off to ensure safer learning (Fig. \ref{fig:exp1detailed} (top)).

We also evaluate the average number of violations for the CPO and TRPO policies after training. We find that the CPO policy continues to maintain fewer violations and is thus safer to roll out for real-world applications than the TRPO policies even though all the policies can perform the task successfully $\ge 95\%$ of the times (Fig. \ref{fig:testbargraph}).

\subsection{Experiment 2: Effect of Boundary Radius}

We evaluate the effect of changing the boundary radius in our constraint formulation on training a CPO policy. We find that a tighter radius takes significantly more samples to train, whereas a relaxed radius is more sample efficient (Fig. \ref{fig:exp234} (left top)). This behavior can also be reinforced by the average number of violations reducing more quickly for a relaxed constrained than a tighter one, although it is obviously lower to begin with (Fig. \ref{fig:exp234} (left bottom)).

\subsection{Experiment 3: Effect of Cost Limits}

We also evaluate how changing the overall cost limit affects the way CPO policies are trained. We see that as the cost limit decreases, the sample efficiency also decreases (Fig. \ref{fig:exp234} (center top)). From the average number of violations plot (Fig. \ref{fig:exp234} (center bottom)), we see that CPO optimizes perfectly for the respective limits and maintains the allowed cost throughout most of the training. 

\subsection{Experiment 4: Effect of Penalty Costs}

We evaluate how changing the scale of penalties impacts the way CPO trains. We linearly scale the cost limits in this case to maintain the same number of allowed constraint violations. We observe that the higher the penalty cost per violation, the longer it takes for the policy to train (Fig. \ref{fig:exp234} (right top)). However, scaling of the penalty costs does not really impact how the average number of violations reduce during training (Fig. \ref{fig:exp234} (right bottom)).

\section{Conclusion}

We explore adding constraints to an object relocation task to potentially enable safe training on real robots. We formulate a constraint that restricts a robot hand's motion to within a boundary when approaching the object. We then learn a policy that uses CPO to optimize the constraint cost. We find that learning to follow the constraints via CPO reduces the average cost during training and roll outs, especially when compared to TRPO. We observe consistency in this result across different constraint boundaries and throughout the training process. We also evaluate the effect of changing the boundary radius, cost limits, and penalty costs on training CPO. We find that smaller constraints and larger penalty costs reduce training efficiency. We conclude that the cylindrical boundary constraint we formulate for the relocation task can help to quickly learn safe motion in training and roll out, and can thus be used for training dexterous manipulation tasks safely on real world robots.

\section{Future Work}
To further investigate the robustness of our boundary constraint and approach, we plan to evaluate our methods on additional dexterous manipulation tasks, such as using a hammer and opening a door. We also plan to formulate a constraint that restricts the motion of the robot after the object has been grasped to further ensure safety throughout the trajectory. Finally, we plan to implement the CPO algorithm on a real robot, such as Shadow Hand and evaluate the effectiveness of our algorithm for real-world applications.



\bibliographystyle{named}
\bibliography{ijcai22}

\pagebreak

\appendix

\section{Deriving the constraints}
\label{appendix:a}

Consider $\textbf{x}_h = (x_h, y_h, z_h)$ to be the initial position of the hand, $\textbf{x}_b = (x_b, y_b, z_b)$ to be the initial position of the object. Any point, $\textbf{p}$ lying on a line passing through these two points can be written using a parameter $t$ as:

\renewcommand{\theequation}{\thesection.\arabic{equation}}

\begin{equation}
\label{eq:parametric}
\textbf{p} = \begin{bmatrix}
x_h + (x_b - x_h)t\\
y_h + (y_b - y_h)t\\
z_h + (z_b - z_h)t
\end{bmatrix}
\end{equation}

The squared distance between any point on this line with parameter $t$ and a point $\textbf{x} = (x, y, z)$ is given by:

\begin{equation}
\label{eq:dist}
\begin{aligned}
    d^2 = [(x_h - x) + (x_b - x_h)t]^2
    \\ + [(y_h - y) + (y_b - y_h)t]^2
    \\ + [(z_h - z) + (z_b - z_h)t]^2
\end{aligned}
\end{equation}

To minimize this distance, we set $\partial (d^2) / \partial t = 0$ and solve for $t$ to get:

\begin{equation}
t = \frac{(\textbf{x} - \textbf{x}_h) \cdot (\textbf{x}_b - \textbf{x}_h)}{|\textbf{x}_b - \textbf{x}_h|^2}
\end{equation}

Plugging $t$ back into Eq. \ref{eq:dist} and simplifying, we get:

\begin{equation}
\label{eq:simplify_dist}
\begin{aligned}
    d^2 = \frac{|\textbf{x}_h - \textbf{x}|^2|\textbf{x}_b - \textbf{x}_h|^2 - [(\textbf{x}_h - \textbf{x}) \cdot (\textbf{x}_b - \textbf{x}_h)]^2}{|\textbf{x}_b - \textbf{x}_h|^2}
\end{aligned}
\end{equation}

Using the vector quadruple product $$(\textbf{A}\times\textbf{B})^2 = \textbf{A}^2\textbf{B}^2 - (\textbf{A} \cdot \textbf{B})^2$$ 

we obtain

$$
d^2 = \frac{|(\textbf{x}_b - \textbf{x}_h) \times (\textbf{x}_h - \textbf{x})|^2}{|\textbf{x}_b - \textbf{x}_h|^2}
$$

Applying the square root on both sides leads us to the following equation:

\begin{equation}
\label{eq:final_dist}
    d = \frac{|(\textbf{x}_b - \textbf{x}_h) \times (\textbf{x}_h - \textbf{x})|}{|\textbf{x}_b - \textbf{x}_h|}
\end{equation}

To define a cylindrical boundary constraint as visualized in Fig. \ref{fig:taskconstraint}, we want the hand position $\textbf{x}$ to be within a fixed distance $r$ of the line joining $\textbf{x}_h$ and $\textbf{x}_b$. This implies that $d \leq r$, or

\begin{equation}
\label{eq:constraint1}
    \frac{|(\textbf{x}_b - \textbf{x}_h) \times (\textbf{x}_h - \textbf{x})|}{|\textbf{x}_b - \textbf{x}_h|} \leq r
\end{equation}

Since we also do not want to consider the entire line joining $\textbf{x}_h$ and $\textbf{x}_b$, but only the line segment joining these two points, from Eq. \ref{eq:parametric}, we can say that $0 \leq t \leq 1$, or

\begin{equation}
\label{eq:constraint2}
    0 \leq  \frac{(\textbf{x} - \textbf{x}_h)\ .\ (\textbf{x}_b - \textbf{x}_h)}{|\textbf{x}_b - \textbf{x}_h|^2} \leq 1
\end{equation}

Thus, Eq. \ref{eq:constraint1} and Eq. \ref{eq:constraint2} form our set of constraints for the objection relocation dexterous manipulation task.

\section{Constraint configurations}
\label{appendix:b}

We use five constraint parameters for our experiments --  boundary radius ($r$), penalty cost ($c$), cost limit ($cl$), minimum $t$ value ($t_{min}$), and maximum $t$ value ($t_{max}$).

For all our experiments, $t_{min}$ is fixed at $-0.1$ and $t_{max}$ is fixed at $1.1$. We relax the original constraint range between $0$ and $1$ (Eq. \ref{eq:constraint2}) for practical purposes.

\subsection{Experiment 1: Evaluating CPO and TRPO}

For the first experiment, we use a fixed penalty cost ($c = 0.01$) and cost limit ($cl = 0.25$), and use three different constraint radii.

\begin{itemize}
    \item $r = 0.1$ for large constraint radius
    \item $r = 0.05$ for medium constraint radius
    \item $r = 0.03$ for small constraint radius
\end{itemize}

\subsection{Experiment 2: Effect of Boundary Radius}

For the second experiment, we use fixed penalty cost ($c = 0.01$) and cost limit ($cl = 0.25$), and use three different constraint radii.

\begin{itemize}
    \item $r = 0.15$ for large constraint radius
    \item $r = 0.05$ for medium constraint radius
    \item $r = 0.03$ for small constraint radius
\end{itemize}

\subsection{Experiment 3: Effect of Cost Limits}

For the third experiment, we use fixed penalty cost ($c = 0.01$) and boundary radius ($r = 0.05$), and use three different cost limits.

\begin{itemize}
    \item $cl = 0.5$ for large cost limit
    \item $cl = 0.25$ for medium cost limit
    \item $cl = 0.1$ for small cost limit
\end{itemize}

\subsection{Experiment 4: Effect of Penalty Costs}

For the fourth experiment, we use a fixed boundary radius ($r = 0.05$), and three different penalty costs. We scale the cost limits linearly with the penalty costs. 

\begin{itemize}
    \item $c = 10, cl = 250$ for large penalty cost
    \item $c = 0.1, cl = 2.5$ for medium penalty cost
    \item $c = 0.01, cl = 0.25$ for small penalty cost
\end{itemize}

\end{document}